\documentclass[10pt,twocolumn,letterpaper]{article}

\usepackage{cvpr}              
\usepackage{multirow}
\definecolor{cvprblue}{rgb}{0.21,0.49,0.74}
\usepackage[pagebackref,breaklinks,colorlinks,allcolors=cvprblue]{hyperref}

\def\paperID{34341} 
\def\confName{CVPR}
\def\confYear{2026}

\title{MHMamba: Multi-Head Mamba for 3D Brain Tumor Segmentation}

\author{
	Hanjun Tao$^{1}$ \quad Hua Wang$^{2}$ \quad Fan Zhang$^{1,*}$\\
	$^{1}$Shandong Technology and Business University, China\\
	$^{2}$Ludong University, China\\
	{\tt\small 2024420192@sdtbu.edu.cn \quad hua.wang@ldu.edu.cn \quad zhangfan@sdtbu.edu.cn}\\
	{\tt\small *Corresponding author}
}

\begin{document}
\maketitle
\begin{abstract}
Brain tumors exhibit high heterogeneity in morphology and multimodal contrast, making manual slice-by-slice de lineation time-consuming and experience-dependent, thus necessitating efficient and stable automated segmentation methods. To address the limitations of CNNs in modeling long-range dependencies, and the heavy computational and memory overhead and inter-block contextual in coherence of Transformers in 3D MRI, this paper proposes Multi-Head Mamba (MHMamba). This method combines a U-shaped architecture with a multi-head state-space model (Mamba), splitting the channel dimension into parallel SSM heads and aggregating them with residuals. This enhances long-range representation and improves the stability of multimodal training while maintaining linear complexity. To further align statistics and enhance lesion response, we designed a channel-space calibration module for multi-head outputs and introduced an adaptive fusion mechanism at skip connections to dynamically connect global semantics with local details, thereby improving boundary consistency and the detection of small-volume lesions. We conducted experiments and ablations on BraTS2021 and BraTS2023. The results showed that MHMamba achieved stable and significant improvements in overall accuracy, boundary smoothness, and sensitivity to tumor core and small-volume enhancement areas, while preserving the linear-complexity advantage of Mamba-based modeling, thus verifying the effectiveness and versatility of the method.

\end{abstract}
    
\section{Introduction}
\label{sec:intro}

Brain tumors, due to their inherent complexity and high variability among patients, are among the deadliest cancers\cite{ostrom2022cbtrus}. They are typically classified as high-grade or low-grade, composed of different tumor subregions, including enhancing tumors, necrotic cores, and surrounding edema\cite{menze2014multimodal}. Magnetic resonance imaging (MRI) is the preferred clinical approach for diagnosing and analyzing brain tumors, and accurate automated segmentation is crucial for developing individualized treatment plans such as radiotherapy planning, preoperative assessment, and efficacy monitoring\cite{litjens2017survey}. However, manually delineating each slice on MRI sequences is time-consuming, labor-intensive, and highly dependent on clinical experience and subjective thresholds, further highlighting the urgent need for efficient and accurate automated segmentation technology.

In recent years, CNNs and Transformers have each had their advantages in medical image segmentation. CNNs rely on local inductive biases, providing strong boundary and texture modeling, but their expansion of the receptive field largely depends on layer stacking and downsampling, resulting in insufficient long-range dependence and a tendency to damage details in small lesions\cite{ronneberger2015u}. While Transformers can establish global relationships, they incur computational and memory costs that increase quadratically with the number of tokens (T), making them insufficient even at the ${128^3}$ patch scale of 3D MRI and more difficult to generalize stably with small datasets\cite{vaswani2017attention}. To reduce complexity, windowing/block attention is a common method, but the fixed or semi-fixed interaction radius has limited adaptability to scale changes and deformations. The Trans-CNN hybrid paradigm attempts to strike a balance between efficiency and context capture\cite{chen2021transunet}. However, it remains limited by multiple factors in multimodal 3D brain tumor scenarios. First, the inconsistent expression domains of convolution and attention in sampling mechanisms and normalization (BN/IN vs. LN) easily lead to feature statistical mismatch and weaken fusion stability. Second, increasing resolution and token number to balance details and context results in a significant increase in memory, bandwidth, and latency. Finally, the superposition of windowed attention and sliding window inference easily causes cross-window context incoherence and makes the predicted probability field unsmooth, thus weakening boundary consistency.

Recently, state-space models (SSMs) have been used to model long sequences with linear complexity, offering both speed and scalability\cite{gu2024mamba}. However, for 3D MRI with its varied morphology and significant modal differences, simple sequential SSMs often suffer from unstable training convergence, insufficient global-local representation capabilities, and inadequate information fusion at skip connections; these issues are particularly pronounced in the detailed characterization of ring enhancement, the boundary between necrosis and living tissue, and small-volume ectopic tumors (ETs).

To overcome these challenges, this paper proposes MHMamba for brain tumor segmentation, combining a U-shaped structure with the Mamba algorithm to model global features of the entire volume at different scales. To enhance long-range dependency modeling capabilities while maintaining linear complexity and improving training stability and adaptability under multimodal conditions, a multi-head Mamba module is designed. Subsequently, a Channel Spatial Calibration (CSCA) module is further designed to reweight the output in terms of channel and spatial dimensions, unifying feature scales and highlighting lesion-related features, thus making the multi-head Mamba output more stable and usable. Furthermore, an Adaptive Gated Fusion (AGF) module is designed at skip connections to dynamically fuse multi-scale features through a learnable adaptive fusion mechanism, balancing details and semantics, and improving boundary consistency and small lesion recognition rate. Systematic experiments and ablation tests were conducted on BraTS2021\cite{baid2021rsna}  and BraTS2023\cite{li2024brain}. The results show that the proposed MHMamba significantly improves overall segmentation accuracy, boundary consistency, and sensitivity to small lesions while maintaining high efficiency, validating the effectiveness and versatility of the method. The contributions of this paper are as follows:

\begin{itemize}
	\item  This paper proposes a multi-head Mamba for 3D MRI of brain tumors. It decomposes channels into parallel SSM heads and aggregates them using residuals, enhancing long-range modeling capabilities while maintaining linear complexity, improving training stability under multimodal conditions, and better adapting to multimodal information.
	\item  To improve the long-range dependency modeling capability and output stability of the multi-head Mamba, this paper introduces a channel space calibration module and an adaptive gated fusion module. The CSCA module calibrates the channel-space of the multi-head Mamba to align feature statistics and scale, while the AGF module adaptively fuses multi-scale features at skip connections to better connect global semantics and local details.
	\item  Experimental results on the BraTS2021 and BraTS2023 brain tumor segmentation tasks show that MHMamba achieves stable improvements in overall segmentation accuracy, boundary consistency, and sensitivity to small lesion detection, especially in tumor core region and boundary prediction, exhibiting higher robustness and accuracy.
	
\end{itemize}

\section{Related works}

\textbf{Brain tumor segmentation methods based on CNN.} In the field of medical image segmentation, FCN\cite{zhou2017deep} transforms the classification network into a structure that can output segmentation maps, providing a starting point for end-to-end training; U-Net\cite{ronneberger2015u} first proposed a symmetrical encoder-decoder U-shaped structure and compensated for the information loss caused by downsampling through skip connections, and quickly became the basic framework for medical segmentation. In response to the special needs of brain tumor tasks, V-Net\cite{milletari2016v} first introduced three-dimensional convolution into medical image segmentation, utilizing the spatial continuity of volume data and stabilizing training through residual connections; 3D U-Net\cite{qamar2020variant} fully utilizes the spatial continuity between slices on this basis and has been widely used in brain tumor segmentation tasks. In order to enhance the spatial consistency of brain tumor segmentation, DenseCRF\cite{cheng2023semantic} is used as a post-processing method in the inference stage to apply boundary constraints, making the segmentation results more consistent with anatomical features; Attention U-Net\cite{oktay2018attention} introduces attention gating on the decoding side to alleviate the confusion between organ and tumor boundaries. Although these improvements have boosted CNN performance, they still fail to overcome the limitation of
local receptive fields in modeling long-range spatial dependencies, and continuous down-sampling weakens edge and detail information. Even with auto-dynamic convolution and location-aware attention\cite{wang2026eeo}, the effective receptive field remains essentially restricted by local convolution operations.

\begin{figure*}[t]  
	\centering
	\includegraphics[width=0.65\textwidth]{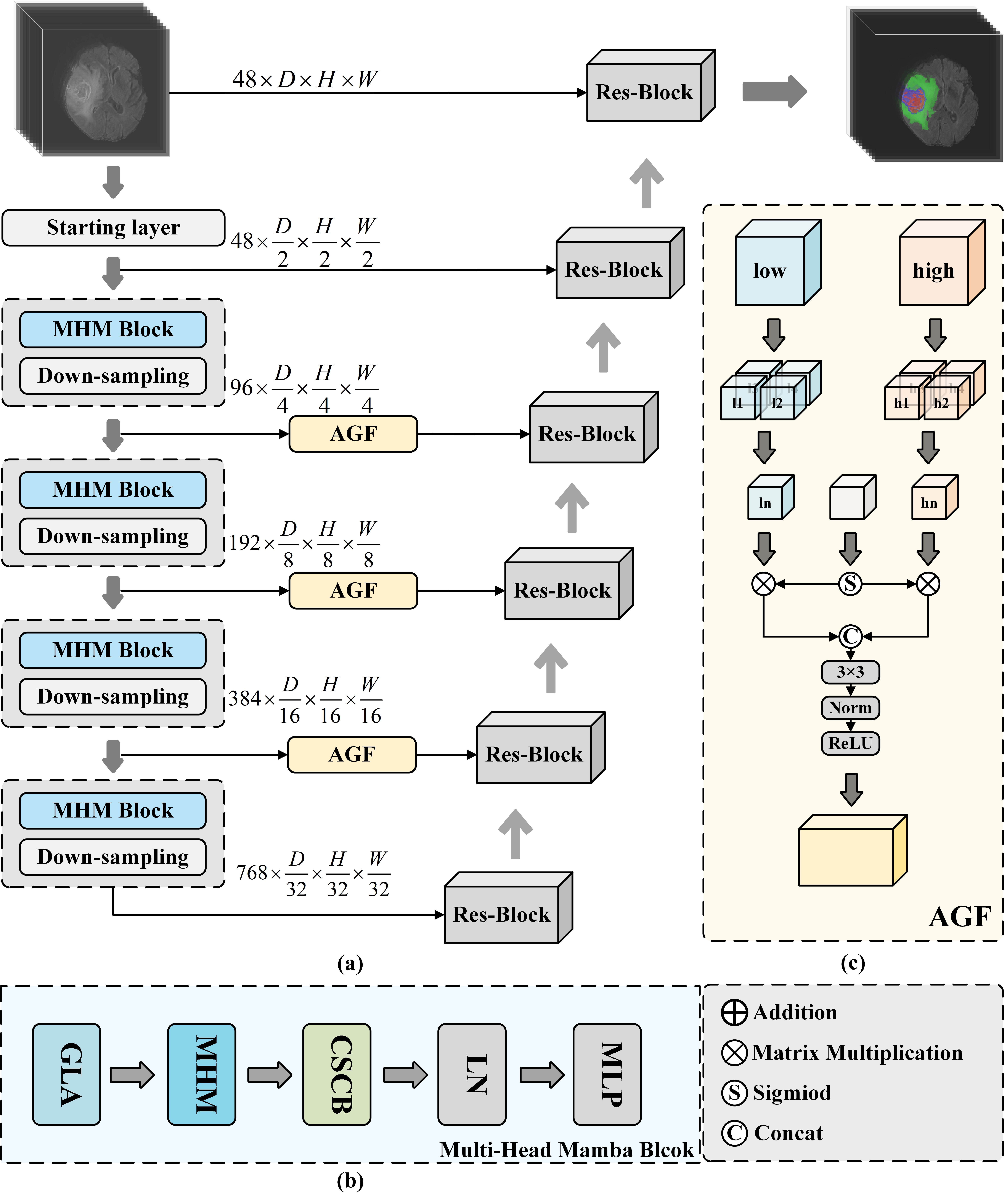}
	\caption{(a) shows the overall structure of the multi-head Mamba model, (b) shows the structure of the Multi-Head Mamba block in the encoding stage, and (c) shows the structure of the adaptive gating fusion module (AGF).}
\end{figure*}

\textbf{Brain tumor segmentation methods based on Transformer.}
With the success of Vision Transformer in natural image tasks, self-attention has been introduced into medical image segmentation to overcome the limited receptive field of convolutional models. Compared with traditional attention mechanisms (e.g., SENet\cite{hu2018squeeze}, CBAM\cite{woo2018cbam}), Transformer naturally models global context and captures long-range dependencies, which is crucial for segmenting structures with complex anatomy. Early work such as TransUNet incorporated ViT into an encoder–decoder framework, but its 2D formulation limited the modeling of 3D spatial continuity. To address this, TransBTS\cite{wang2021transbts} combined 3D CNNs with Transformer blocks for local–global feature extraction, improving enhancing tumor (ET) segmentation. Building on TransBTS, TMA-TransBTS introduces 3D multi-scale self-attention and cross-attention to strengthen volumetric context modeling and long-range dependency capture\cite{huang2025multi}. Swin Transformer further achieves a balance between local interactions and global modeling via its hierarchical shifted-window design, while TransAttUNet\cite{chen2023transattunet} enhances multi-scale feature propagation to mitigate edema (ED) feature decay\cite{zhang2024cf,zhang2023multi}. Meanwhile, constrained polynomial approximation and node construction have also been used to enhance representation flexibility in dense prediction tasks\cite{wang2024computing}.

Recent studies have also improved Transformer-based segmentation through channel–spatial attention and lightweight multi-scale fusion. VcaNet integrates a ViT backbone with fusion channel–spatial attention to better preserve 3D tumor boundaries\cite{pan2025vcanet}, and MCSLF-Net employs multi-level channel–spatial attention with a light-weight scale-fusion Transformer to balance accuracy and efficiency\cite{zhou2025multi}. Overall, Transformer architectures effectively capture long-range dependencies and complex tumor morphology in multimodal 3D MRI.

However, the quadratic complexity of standard self-attention ($O(N^{2})$) restricts the resolution, batch size, and depth feasible for 3D MRI. Block attention alleviates computation by limiting interactions locally, but cross-block context propagation remains weak; clipping and stitching during inference additionally cause probability-field discontinuity and boundary inconsistency\cite{isensee2021nnu}. Moreover, Transformers typically use layer normalization (LN), whereas CNN branches rely on batch normalization (BN) or instance normalization (IN), and such statistical mismatch may lead to inconsistent features and discontinuities at block boundaries.

\textbf{Brain tumor segmentation methods based on Mamba.} Under 3D MRI conditions, it is crucial to preserve high-resolution details while capturing a large contextual range. State space models (SSMs) can model long-range dependencies with near-linear complexity, and Mamba realizes this advantage through selective state updates and efficient parallel scanning. VMamba\cite{liu2024vmamba} replaces self-attention with visual SSM blocks in the encoder to obtain a larger receptive field with linear complexity, while retaining convolution and skip connections in the decoder to preserve details and efficiency. U-Mamba\cite{ma2024u} mainly places Mamba blocks at high-level and bottleneck stages, keeps convolutional layers at shallow stages to protect boundaries and textures, and aligns bidirectional features in the decoder to balance global context and fine details. SegMamba\cite{xing2024segmamba} adopts a multi-scale hierarchical SSM design, where deeper layers focus on global relationships and shallower layers retain structural details, and further introduces boundary alignment constraints during clipping–stitching inference to reduce seam artifacts.

Recent advances have extended Mamba/SSM architectures in various directions. For example, LS3M employs learnable sorting to better handle incomplete multimodal MRI\cite{zhang2025incomplete}, while DRBD-Mamba enhances long-range dependency modeling through a dual-resolution bidirectional design\cite{ali2025drbd}. In addition, consistency-driven SSMs introduce cross-modality consistency regularization to improve robustness in multimodal integration\cite{liu2025consistency}. Building upon these ideas, the Multi-Head Mamba proposed in this work performs channel decomposition and parallel SSM processing, enhancing the capacity and directional sensitivity of global representations while maintaining near-linear complexity, thereby better meeting the 3D segmentation needs of multimodal and morphologically diverse brain tumors.

\section{Methods}
Multi-head Mamba employs a U-shaped encoder--decoder architecture, primarily composed of three parts: 1) a 3D feature encoder based on multi-head Mamba blocks, used to model global information and calibrate channel and spatial features at different scales; 2) a 3D decoder based on convolutional layers, which recovers spatial resolution and predicts segmentation results through upsampling; and 3) skip connections based on adaptive fusion, used to enhance multi-scale features. Figure~1(a) shows the overall architecture of MHMamba.

\subsection{Multi-head Mamba module}

Brain tumor segmentation requires both sufficient global context and the maintenance of high-resolution details within a small ET volume. Directly using a Transformer for global modeling typically relies on a large downsampling ratio to shorten the sequence length, which reduces complexity but may result in the loss of multi-scale information at the encoding stage. To achieve linear global modeling capability while maintaining high resolution, we designed an MHMamba block, introduced into each encoding scale.
The encoder consists of a stem initial layer and multiple MHMamba blocks stacked sequentially. Specifically, given a 3D input volume
${\rm{X}} \in {{\rm{R}}^{B \times {C_{{\rm{in}}}} \times D \times H \times W}}$,
the input is first projected into the $C$-dimensional embedding space through the initial layer. The initial layer uses a depthwise separable convolution ($7 \times 7 \times 7$ depthwise, stride $2$, padding $3$), resulting in the first-scale feature ${F_{\rm{0}}} \in {{\rm{R}}^{B \times 48 \times \frac{D}{2}{\rm{ }} \times \frac{H}{2}{\rm{ }} \times \frac{W}{2}}}$.
Then, for stages $i=1,2,3,4$, $F_0$ passes through the MHM blocks sequentially and is downsampled at the end of each stage using a $3 \times 3 \times 3$ convolution with stride $2$. The spatial size is halved, and the number of channels increases from $C_{i-1}$ to $C_i$. This four-scale representation, from shallow to deep, expands the contextual coverage while retaining fine-grained information useful for small lesions.

As shown in Figure~1(b), in the MHM block, local boundary enhancement is first performed using Gated Local Aggregation (GLA), followed by global sequence modeling with linear complexity using Multi-Head Mamba (MHM). Then, Channel-Spatial Calibration Attention (CSCA) performs reweighting and statistical alignment on both channel and spatial paths. Finally, LayerNorm and MLP are used to complete channel mixing and residual shaping. The following sections will detail the Global-Local Aggregation module, the Multi-Head Mamba module, and the Channel-Spatial Collaborative Attention module.

\subsubsection{GLA}

Gated Local Aggregation serves as the starting point for the MHM module. It employs a parallel structure to enhance the geometric feature representation of tumor boundaries, providing rich boundary information for global MHM modeling and enhancing sensitivity to tumor edges.
\begin{equation}
	F_{\mathrm{edge}}=\mathrm{Sobel3D}(F)
	\label{eq:gla_edge}
\end{equation}
\begin{equation}
	F_{\mathrm{detail}}=\mathrm{Conv}(\mathrm{ReLU}(\mathrm{IN}(F)))
	\label{eq:gla_detail}
\end{equation}
\begin{equation}
	F_{\mathrm{GLA}}=\alpha \cdot F_{\mathrm{edge}}+\beta \cdot F_{\mathrm{detail}}
	\label{eq:gla_out}
\end{equation}

Where Sobel3D is the 3D Sobel operator used to extract boundary features, IN denotes instance normalization, and $\alpha$ and $\beta$ are learnable weight parameters. This module provides rich local geometric features for subsequent global modeling by enhancing boundary and detail information.

\begin{figure*}[t]
	\centering
	\includegraphics[width=0.7\textwidth]{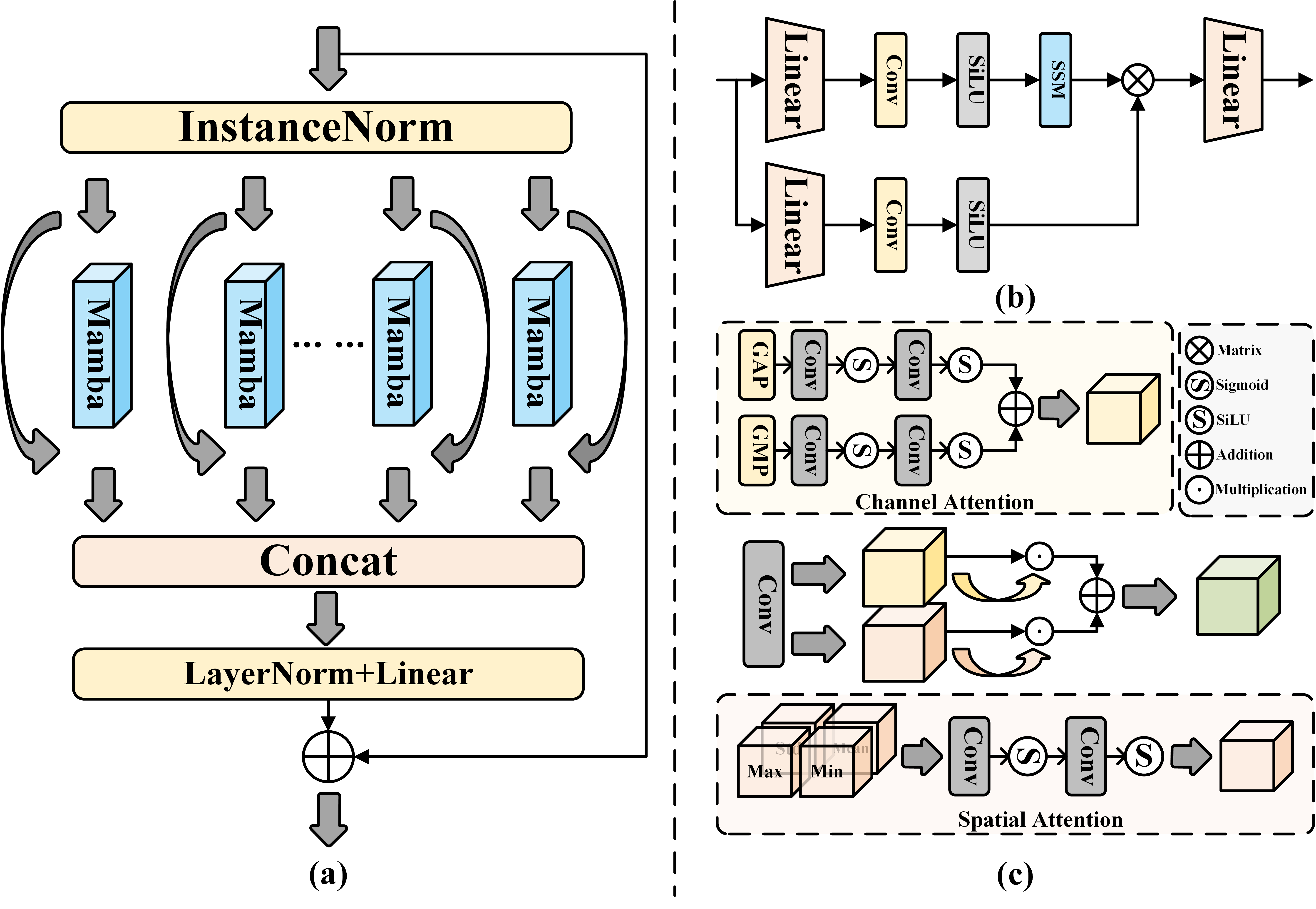}
	\caption{(a) shows the specific structure of the Multi-head Mamba block, and (b) shows the architecture of the Mamba block. We adopt a dual-channel strategy, introducing a symmetric token mixing path without SSM to capture cross-region dependencies while replacing causal convolutions with regular convolutions. Finally, we enhance the modeling of the global context by achieving adaptive fusion through gated weighting. (c) shows the structure of the Channel-Spatial Calibration module.}
\end{figure*}

 \subsubsection{Multi-head Mamba}

Three-dimensional brain tumor MRI exhibits structural relationships with varying orientations and modality-specific responses. Modeling long-range dependencies using a single path can easily mix information from different modalities, leading to ``averaging'' of the representation. In the early stages of training, excessive global branching may also suppress local details. Therefore, we propose Multi-head Mamba, which employs a multi-head approach as shown in Figure~2(a). This improves selectivity and capacity without altering the main complexity, thus balancing global context and stable convergence under 3D conditions. The specific process is as follows.
\begin{equation}
	F_{\mathrm{norm}} = \mathrm{LN}(F_{\mathrm{GLA}})
	\label{eq:mhm_norm}
\end{equation}

\begin{equation}
	\left[F^{1}, F^{2}, \ldots, F^{N_h}\right] = \mathrm{Split}(F_{\mathrm{norm}}, N_h)
	\label{eq:mhm_split}
\end{equation}

This partition enables different heads to learn complementary long-range dependency patterns from the joint multimodal input \cite{ENCODER,REFINE}. We do not impose a fixed modality-to-head assignment; instead, head specialization is learned automatically from the data, which is broadly consistent with efforts on fine-grained semantic parsing and reversible alignment in heterogeneous multimodal settings \cite{FineCIR,xiaoreversible}.

\begin{equation}
	F_{\mathrm{flat}}^{j} = \mathrm{Flatten}(F^{j}) \in \mathbb{R}^{B \times N \times \frac{C}{N_h}}
	\label{eq:mhm_flat}
\end{equation}

Each head first undergoes serialization, converting the 3D spatial features into a sequence format. Where $N = D \times H \times W$ represents the total number of spatial locations. This step transforms the 3D volumetric data into a format suitable for sequence modeling while preserving the relative relationships between spatial locations.
\begin{equation}
	h_t^{j} = \bar{A}_t^{j} h_{t-1}^{j} + \bar{B}_t^{j} x_t^{j}
	\label{eq:mhm_state}
\end{equation}

\begin{equation}
	y_t^{j} = C_t^{j} h_t^{j} + D_t^{j} x_t^{j}
	\label{eq:mhm_output}
\end{equation}

The parameters are dynamically generated through an input-dependent selective mechanism, enabling the model to adaptively adjust the state transition process based on the input content. Specifically, in brain tumor images, the model can allocate more attentional resources to enhance tumor regions while employing simpler state transitions for normal brain tissue regions. This dynamic adjustment capability significantly improves the model's computational efficiency and feature selectivity. The outputs of all heads are fused through concatenation and projection, and then added to the input features along with learnable residual weights. This design ensures the effective integration of global contextual information and local features, providing rich semantic information for subsequent feature calibration.

\begin{equation}
	F_{\mathrm{concat}} = \mathrm{Concat}(F_{\mathrm{ssm}}^{1}, F_{\mathrm{ssm}}^{2}, \ldots, F_{\mathrm{ssm}}^{N_h})
	\label{eq:mhm_concat}
\end{equation}

\begin{equation}
	F_{\mathrm{MHM}} = W_p \cdot F_{\mathrm{concat}} + \delta \cdot F_{\mathrm{GLA}}
	\label{eq:mhm_final}
\end{equation}

Where $W_p$ is a $1 \times 1 \times 1$ projective convolution kernel and $\delta$ is a learnable residual weight.

The innovation of the Multihead Mamba module lies in its successful implementation of global modeling with parallel linear complexity, a feature that makes it particularly suitable for processing high-resolution 3D medical images. Compared to traditional self-attention mechanisms, the Multihead Mamba module offers significant advantages in computational efficiency and memory usage, while maintaining the model’s ability to capture long-range dependencies.
\subsubsection{CSCA}
The Channel-Space Calibration Module is a key component in MHMamba for global feature calibration, as shown in Figure~2(c). Its goal is to simultaneously perform statistical alignment of the channel and spatial dimensions after multi-head Mamba and enhance tumor-related responses. The module employs a dual-path design. In the channel attention path, we utilize a dual-path design of global average pooling and global max pooling, generating channel weights through a multilayer perceptron to highlight the contributions of important feature channels. Its calculation can be expressed as
\begin{equation}
	\begin{array}{l}
		{F_{c}} = \sigma \big({\rm{MLP}}({\rm{GAP}}({F_{{\rm{MHM}}}})) + \\
		{\rm{MLP}}({\rm{GMP}}({F_{{\rm{MHM}}}}))\big) \cdot F_{{\rm{MHM}}}^c
	\end{array}
\end{equation}
Where $\sigma$ is the sigmoid function.

In the spatial attention path, we innovatively integrate four spatial statistics (mean, standard deviation, maximum, and minimum) and generate a spatial weight map through splicing and activation to enhance the feature response of key tumor regions. The calculation process is as follows
\begin{equation}
	\begin{aligned}
		F_s
		&= \sigma \Big(
		\operatorname{Concat}\big(
		\operatorname{Mean}(F_{\mathrm{MHM}}),\,
		\operatorname{Std}(F_{\mathrm{MHM}}), \\
		&\qquad\qquad
		\operatorname{Max}(F_{\mathrm{MHM}}),\,
		\operatorname{Min}(F_{\mathrm{MHM}})
		\big)
		\Big)\cdot F_{\mathrm{MHM}}^{s}
	\end{aligned}
\end{equation}

These four statistics summarize overall response, salient peaks, heterogeneity, and suppression, respectively, and provide a more stable spatial calibration signal for heterogeneous tumor regions \cite{ReTrack,OFFSET}. A gating coordination mechanism is then introduced to adaptively fuse the channel and spatial branches, while the residual connection preserves the original feature response and improves optimization stability; related uncertainty-aware and discrimination-aware designs have also been explored in broader retrieval settings \cite{HUD,INTENT}.
\begin{equation}
\begin{array}{l}
	{F_{{\rm{CSCA}}}} = \lambda \cdot {F_c} + (1 - \lambda) \cdot {F_s} + {F_{{\rm{MHM}}}} \\
	\lambda = \sigma ({w_{{\rm{gate}}}} \cdot [{F_c},{F_s}])
\end{array}
\end{equation}

\begin{equation}
	F_{\mathrm{out}} = \operatorname{MLP}\!\left(\operatorname{LN}\!\left(F_{\mathrm{CSCA}}\right)\right)
\end{equation}
where $\lambda$ is the gating weight. 
\subsection{Adaptive Gated Fusion (AGF)}
In decoder design, we face a key challenge: how to effectively fuse multi-scale features from the encoder? Traditional skip connections typically employ simple concatenation or addition operations, but this approach is clearly insufficient for brain tumor segmentation tasks. Features at different levels possess vastly different semantic information and spatial details, and direct fusion can easily lead to feature conflicts, especially in the tumor boundary region, resulting in inconsistent segmentation results. To address this, we propose an Adaptive Gated Fusion module. The core idea of this module is to establish an adaptive feature fusion mechanism. As shown in Figure~1(c), the AGF module employs a staged processing strategy, achieving optimal fusion of encoder and decoder features at each decoding level.
First, the input features are divided into four subgroups along the channel dimension. For each feature subgroup, the AGF module learns independent fusion weights. Based on the learned weights, each subgroup undergoes adaptive fusion. Finally, all fused subgroups are integrated through a convolution layer. The specific process is as follows:

\begin{equation}
	\big[\mathbf{F}^{1}_{\mathrm{enc}},\, \mathbf{F}^{2}_{\mathrm{enc}},\, \mathbf{F}^{3}_{\mathrm{enc}},\, \mathbf{F}^{4}_{\mathrm{enc}}\big]
	= \operatorname{Split}\!\big(\mathbf{F}_{\mathrm{enc}},\, 4\big)
\end{equation}
\begin{equation}
	\big[\mathbf{F}^{1}_{\mathrm{dec}},\, \mathbf{F}^{2}_{\mathrm{dec}},\, \mathbf{F}^{3}_{\mathrm{dec}},\, \mathbf{F}^{4}_{\mathrm{dec}}\big]
	= \operatorname{Split}\!\big(\mathbf{F}_{\mathrm{dec}},\, 4\big)
\end{equation}
\begin{equation}
	\begin{array}{l}
		F_{{\rm{fused}}}^k = {\delta _k} \cdot F_{{\rm{enc}}}^k + (1 - {\delta _{k}}) \cdot F_{{\rm{dec}}}^k \\
		{\delta _k} = \sigma ({w_k} \cdot [F_{enc}^k,F_{dec}^k])
	\end{array}
\end{equation}
\begin{equation}
	{F_{{\rm{out}}}} = {W_f} \cdot {\rm{Concat}}(F_{{\rm{fused}}}^1,F_{{\rm{fused}}}^2,F_{{\rm{fused}}}^3,F_{{\rm{fused}}}^4)
\end{equation}
Where $\sigma$ is the sigmoid function and $w_k$ is a learnable convolution kernel.

Conceptually, the proposed AGF is related to dual-branch feature fusion methods based on unfolded optimization~\cite{zhang2026decoding}, but our design uses a lightweight one-shot gating strategy tailored for 3D encoder-decoder skip fusion. CSCA performs post-aggregation calibration on encoder features, whereas AGF resolves cross-scale conflicts during skip fusion between boundary detail and high-level semantics. Therefore, the two modules target different bottlenecks and are complementary rather than redundant.

\begin{table*}[htbp]
	\centering
	\setlength{\tabcolsep}{3pt}
	\caption{Comparative experiments on the BraTS2021 and BraTS2023 datasets}
	\begin{tabular}{@{}l*{16}{c}@{}}
		\toprule
		\multirow{3}{*}{\textbf{Method}} 
		& \multicolumn{8}{c}{\textbf{BraTS2021}} 
		& \multicolumn{8}{c}{\textbf{BraTS2023}} \\
		\cmidrule(lr){2-9} \cmidrule(lr){10-17}
		& \multicolumn{2}{c}{WT} & \multicolumn{2}{c}{TC} & \multicolumn{2}{c}{ET} & \multicolumn{2}{c}{Avg}
		& \multicolumn{2}{c}{WT} & \multicolumn{2}{c}{TC} & \multicolumn{2}{c}{ET} & \multicolumn{2}{c}{Avg} \\
		\cmidrule(lr){2-3} \cmidrule(lr){4-5} \cmidrule(lr){6-7} \cmidrule(lr){8-9}
		\cmidrule(lr){10-11} \cmidrule(lr){12-13} \cmidrule(lr){14-15} \cmidrule(lr){16-17}
		& Dice & HD & Dice & HD & Dice & HD & Dice & HD
		& Dice & HD & Dice & HD & Dice & HD & Dice & HD \\
		\midrule
		nnformer\cite{zhou2021nnformer} 
		& 91.63 & 6.66 & 88.62 & 6.53 & 86.21 & 4.95 & 88.60 & 6.04
		& 91.01 & 4.95 & 86.08 & 5.30 & 79.29 & 6.34 & 85.74 & 5.25 \\
		
		VcaNet\cite{pan2025vcanet}  
		& 90.33 & 6.32 & 86.52 & 6.51 & \underline{87.55} & 4.33 & 88.10 & 5.72
		& 92.77 & 6.16 & \textbf{91.20} & 5.83 & \underline{86.57} & 10.96 & \underline{90.18} & 7.65 \\
		
		nnU-Net\cite{isensee2021nnu}   
		& 92.40 & 6.63 & 88.62 & 6.04 & 87.23 & 4.07 & 89.40 & 5.58
		& 92.95 & 6.87 & 90.24 & 4.85 & 83.23 & 4.69 & 88.81 & 5.47 \\
		
		LightUNet\cite{zhang2024light} 
		& 92.38 & 5.95 & 88.96 & 6.00 & \textbf{88.38} & 4.37 & 89.90 & 5.44
		& 78.16 & 5.00 & 85.42 & 4.96 & 84.07 & \underline{3.93} & 82.55 & 4.63 \\
		
		SegMamba\cite{xing2024segmamba}    
		& \underline{92.92} & 4.21 & \underline{91.46} & 4.32 & 86.58 & 4.02 & \underline{90.32} & 4.18
		& \underline{93.42} & \underline{3.67} & 90.05 & 4.24 & 84.66 & 4.85 & 89.38 & 4.25 \\	
		
		SegMamba-V2\cite{xing2025segmamba}  
		& 92.59 & \underline{4.19} & 89.27 & \underline{3.54} & 85.57 & \underline{3.71} & 89.14 & \underline{3.80}
		& 93.15 & 3.71 & 90.16 & \underline{4.02} & \textbf{86.64} & \textbf{3.56} & 89.98 & \underline{3.76} \\
		
		\textbf{Ours} 
		& \textbf{93.54} & \textbf{3.74} & \textbf{92.23} & \textbf{2.89} & 87.28 & \textbf{3.50} & \textbf{91.02} & \textbf{3.38}
		& \textbf{93.87} & \textbf{3.30} & \underline{91.10} & \textbf{3.49} & 85.72 & 4.23 & \textbf{90.23} & \textbf{3.67} \\
		\bottomrule
	\end{tabular}
	\label{tab:comparative_brats}
\end{table*}

\section{Experiments}
\subsection{Datasets}

We systematically evaluated the proposed Multi-Head Mamba network on the widely adopted BraTS2021 and BraTS2023 datasets for brain tumor segmentation. Each case contains four MRI modalities, namely T1, T1ce, T2, and FLAIR, which are concatenated as a 4-channel input for all methods. BraTS2021 provides 1,251 training samples and 219 validation cases, while BraTS2023 contains 1,534 training samples. All data underwent standardized preprocessing, including multimodal registration, skull stripping, and isotropic resampling, resulting in a unified size of $240\times240\times155$. The annotations include three tumor sub-regions, namely enhancing tumor (ET), tumor core (TC), and whole tumor (WT).

We used only the official training set for model development and created an internal 70\%/10\%/20\% split for training, validation, and testing, respectively. The random seed was fixed to ensure reproducibility, and the validation set mentioned in this paper refers to the internal 10\% subset.

\subsection{Experimental setup and metrics}

All experiments were implemented in PyTorch 2.1.2 and conducted on a single NVIDIA RTX 3090 GPU. All models were trained from random initialization for 300 epochs with batch size 1, initial learning rate 0.001, weight decay 0.00001, and polynomial learning-rate decay. Here, until convergence only indicates metric saturation within this fixed budget rather than different stopping rules.

To ensure fair comparison, all baseline models were evaluated within a unified SegMamba framework using identical preprocessing, data augmentation, training schedule, optimizer, loss, and sliding-window inference, rather than adopting additional progressive-learning or sample-prioritization strategies from broader vision tasks \cite{HABIT,xiao2026points}. The data augmentation included brightness adjustment, gamma correction, rotation, scaling, mirror flipping, and elastic deformation. During training, $128\times128\times128$ 3D patches were randomly cropped from the original $240\times240\times155$ MRI volumes. Following SegMamba, we used an equally weighted sum of Dice loss and cross-entropy loss. Inference was performed using a unified sliding-window strategy without test-time augmentation or post-processing.

We employed the Dice coefficient to evaluate region-overlap accuracy and the 95\% Hausdorff Distance (HD95) to assess boundary alignment precision.

\subsection{Main results}

We compared Multi-Head Mamba with six strong baseline methods on BraTS2021 and BraTS2023. Since preprocessing, patch sampling, and sliding-window inference can substantially affect 3D segmentation performance, all methods were evaluated under one unified pipeline rather than mixed training recipes. When an official implementation was not directly plug-and-play, we reproduced the method according to its paper setting within the same framework. As shown in Table~1, our method achieves the best or near-best performance on the overall metrics across the two benchmarks. Figure~3 further shows that our predictions are typically less fragmented and exhibit tighter boundaries in challenging heterogeneous regions.

All results in Table~1 were obtained under the same training and inference protocol and averaged over multiple runs.

\begin{figure*}[t]  
	\centering
	\includegraphics[width=0.8\textwidth]{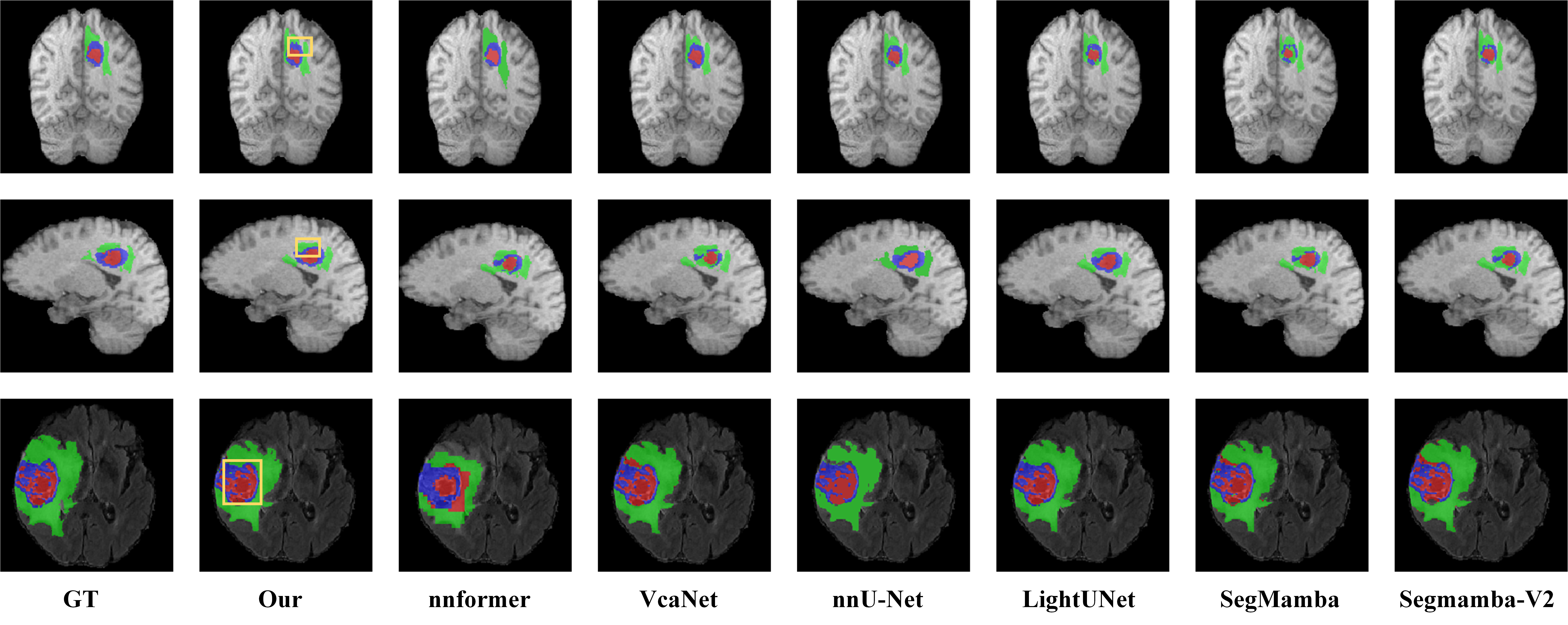}
	\caption{Qualitative visualization results of the BraTS2021 dataset. Key areas marked by yellow boxes demonstrate the significant advantages of our proposed method over other comparative methods in these challenging tumor regions. The figure uses three colors to represent different tumor sub-regions: green represents the whole tumor (WT) region, blue represents the enhanced tumor (ET) region, and red corresponds to the tumor core (TC) region.}
\end{figure*}
\subsection{Ablation experiments}
This section conducts systematic ablation studies to analyze the impact of key components in MHMamba on its segmentation performance. We first evaluate the individual contributions of three core modules, the Multi-Head Mamba module, the Channel-Spatial Calibration module, and the Adaptive Gated Fusion module. Additionally, we investigate how the number of heads in the Multi-Head Mamba module affects the model's performance to determine the optimal architecture design parameters.
\begin{table}[htbp]
	\centering
	\setlength{\tabcolsep}{3pt} 
	\caption{Ablation experiments on the BraTS2021 dataset}
	\begin{tabular}{@{}l*{8}{c}@{}}
		\toprule
		\multirow{2}{*}{\textbf{Method}} & \multicolumn{2}{c}{WT} & \multicolumn{2}{c}{TC} & \multicolumn{2}{c}{ET} & \multicolumn{2}{c}{Avg} \\
		\cmidrule(lr){2-3} \cmidrule(lr){4-5} \cmidrule(lr){6-7} \cmidrule(lr){8-9}
		& Dice & HD & Dice & HD & Dice & HD & Dice & HD \\
		\midrule
		base		& 92.92 & 4.21 & 91.46 & 4.32 & 86.58 & 4.02 & 90.32 & 4.18 \\
		+CSCA 		& 93.42 & 3.33 & 92.03 & 3.57 & 87.04 & 3.53 & 90.83 & 3.47 \\
		+AGF   		& 93.14 & 3.83 & 92.00 & 3.40 & 86.84 & 3.79 & 90.66 & 3.67 \\
		+MHM		& 93.13 & 3.91 & 91.81 & 3.92 & 86.64 & 4.00 & 90.62 & 3.94 \\
		\bfseries ours & \bfseries 93.54 & \bfseries 3.74 & \bfseries 92.23 & \bfseries 2.89 & \bfseries 87.28 & \bfseries 3.50 & \bfseries 91.02 & \bfseries 3.38 \\
		\bottomrule
	\end{tabular}
	
	\label{tab:ablation_brats2021}
\end{table}

\begin{table}[htbp]
	\centering
	\setlength{\tabcolsep}{3pt} 
	\caption{Ablation experiments on the BraTS2023 dataset}
	\begin{tabular}{@{}l*{8}{c}@{}}
		\toprule
		\multirow{2}{*}{\textbf{Method}} & \multicolumn{2}{c}{WT} & \multicolumn{2}{c}{TC} & \multicolumn{2}{c}{ET} & \multicolumn{2}{c}{Avg} \\
		\cmidrule(lr){2-3} \cmidrule(lr){4-5} \cmidrule(lr){6-7} \cmidrule(lr){8-9}
		& Dice & HD & Dice & HD & Dice & HD & Dice & HD \\
		\midrule
		base		& 93.42 & 3.67 & 90.05 & 4.24 & 84.66 & 4.85 & 89.38 & 4.25 \\
		+CSCA 		& 93.61 & 3.61 & 90.77 & 4.03 & 85.10 & 4.79 & 89.83 & 4.14 \\
		+AGF   		& 93.59 & 3.58 & 90.88 & 4.00 & 84.69 & 4.80 & 89.72 & 4.13 \\
		+MHM        & 93.65 & 3.49 & 90.74 & 3.98 & 84.81 & 4.69 & 89.73 & 4.05 \\
		\bfseries ours & \bfseries 93.87 &\bfseries 3.30  & \bfseries 91.10 & \bfseries 3.49 & \bfseries 85.72 & \bfseries 4.23 & \bfseries 90.23 & \bfseries 3.67 \\
		\bottomrule
	\end{tabular}
	
	\label{tab:ablation_brats2023}
\end{table}

\textbf{Role analysis of components.} Based on the Segmamba baseline model, systematic ablative experiments were conducted on the BraTS2021 and BraTS2023 datasets. By progressively integrating the multi-head Mamba module, channel-spatial alignment module, and adaptive gated fusion module, the model demonstrated consistent performance improvements across the three critical tumor sub-regions. As evidenced by the results presented in Tables 2 and 3, the effective functional complementarity and synergistic interactions among these components led to marked improvements in both Dice coefficients and HD95 distances for the WT, TC, and ET regions. Furthermore, the proposed architecture significantly enhanced the model's capacity for feature representation of complex tumor structures and boundary delineation accuracy.

\textbf{Impact analysis of multi-head configuration on model performance.} The introduction of multi-head mechanism aims to enhance the model's capability to capture features from multimodal MRI data and complex tumor morphology. By deploying multiple feature processing heads in parallel, the model can focus on different feature dimensions such as boundary details, long-range dependencies, and modality-specific characteristics. Ablation studies by adjusting the number of heads validate the effectiveness of this design in improving feature diversity representation and determine the optimal balance between computational efficiency and performance. Systematic experiments were conducted on the BraTS2021 dataset. As shown in Table 4, when the number of heads gradually increases from 2 to 8, the model performance demonstrates a trend of initial improvement followed by degradation. Experimental results indicate that the configuration with 4 heads achieves the optimal balance between model representational capacity and computational complexity. This configuration enables different heads to specialize in extracting boundary features, texture features, semantic features, and modality-specific features, thereby achieving comprehensive modeling of multi-dimensional characteristics in brain tumors.
The degradation at $N=8$ is mainly caused by over-partitioning, which narrows each head and makes optimization noisier under 3D training with batch size 1.
\begin{table}[htbp]
	\centering
	\setlength{\tabcolsep}{3pt} 
	\caption{Ablation experiments on the BraTS2021 dataset}
	\begin{tabular}{@{}l*{8}{c}@{}}
		\toprule
		\multirow{2}{*}{\textbf{Head}} & \multicolumn{2}{c}{WT} & \multicolumn{2}{c}{TC} & \multicolumn{2}{c}{ET} & \multicolumn{2}{c}{Avg} \\
		\cmidrule(lr){2-3} \cmidrule(lr){4-5} \cmidrule(lr){6-7} \cmidrule(lr){8-9}
		& Dice & HD & Dice & HD & Dice & HD & Dice & HD \\
		\midrule
		N=2		& 92.99 & 4.17 & 91.73 & 3.98 & 86.61 & 3.97 & 90.44 & 4.04 \\
		\bfseries N=4 & \bfseries 93.54 & \bfseries 3.74 & \bfseries 92.23 & \bfseries 2.89 & \bfseries 87.28 & \bfseries 3.50 & \bfseries 91.02 & \bfseries 3.38 \\
		N=8		& 93.24 & 3.97 & 91.55 & 4.19 & 86.67 & 4.00 & 90.48 & 4.05 \\
		\bottomrule
	\end{tabular}
	
	\label{tab:ablation_brats2023}
\end{table}

\section{Conclusions}

This paper addresses the challenge of balancing global context modeling and local detail preservation in brain tumor segmentation by proposing MHMamba, a segmentation network based on a multi-head state-space model. By integrating the multi-head Mamba module, the channel-spatial calibration module, and the adaptive gated fusion module, the proposed method improves long-range dependency modeling, feature calibration, and multi-scale feature fusion. Experiments on BraTS2021 and BraTS2023 show improved segmentation accuracy and boundary consistency, especially for tumor core regions and small lesions.
\newpage
\section*{Acknowledgments}
This work was supported in part by the following: the National Natural Science Foundation of China under Grant Nos. U24A20219, 62272281, U24A20328, 62576193, the Yantai Natural Science Foundation under Grant No. 2024JCYJ034, and the Youth Innovation Technology Project of Higher School in Shandong Province under Grant No. 2023KJ212.

{
    \small
    \bibliographystyle{ieeenat_fullname}
    \bibliography{main}

@String(ECCV= {Eur. Conf. Comput. Vis.})

@String(AAAI = {AAAI})

@String(ECCV  = {ECCV})

@article{ostrom2022cbtrus,
	title={CBTRUS statistical report: primary brain and other central nervous system tumors diagnosed in the United States in 2015--2019},
	author={Ostrom, Quinn T and Price, Mackenzie and Neff, Corey and Cioffi, Gino and Waite, Kristin A and Kruchko, Carol and Barnholtz-Sloan, Jill S},
	journal={Neuro-oncology},
	volume={24},
	number={Supplement\_5},
	pages={v1--v95},
	year={2022},
	publisher={Oxford University Press US}
}

@article{menze2014multimodal,
	title={The multimodal brain tumor image segmentation benchmark (BRATS)},
	author={Menze, Bjoern H and Jakab, Andras and Bauer, Stefan and Kalpathy-Cramer, Jayashree and Farahani, Keyvan and Kirby, Justin and Burren, Yuliya and Porz, Nicole and Slotboom, Johannes and Wiest, Roland and others},
	journal={IEEE transactions on medical imaging},
	volume={34},
	number={10},
	pages={1993--2024},
	year={2014},
	publisher={IEEE}
}

@article{litjens2017survey,
	title={A survey on deep learning in medical image analysis},
	author={Litjens, Geert and Kooi, Thijs and Bejnordi, Babak Ehteshami and Setio, Arnaud A.A. and Ciompi, Francesco and Ghafoorian, Mohsen and van der Laak, Jeroen A.W.M. and van Ginneken, Bram and S{\'a}nchez, Clara I.},
	journal={Medical Image Analysis},
	volume={42},
	pages={60--88},
	year={2017},
	publisher={Elsevier}
}

@inproceedings{ronneberger2015u,
  title={U-net: Convolutional networks for biomedical image segmentation},
  author={Ronneberger, Olaf and Fischer, Philipp and Brox, Thomas},
  booktitle={International Conference on Medical image computing and computer-assisted intervention},
  pages={234--241},
  year={2015},
  organization={Springer}
}

@article{chen2021transunet,
	title={Transunet: Transformers make strong encoders for medical image segmentation},
	author={Chen, Jieneng and Lu, Yongyi and Yu, Qihang and Luo, Xiangde and Adeli, Ehsan and Wang, Yan and Lu, Le and Yuille, Alan L and Zhou, Yuyin},
	journal={arXiv preprint arXiv:2102.04306},
	year={2021}
}

@article{vaswani2017attention,
	title={Attention is all you need},
	author={Vaswani, Ashish and Shazeer, Noam and Parmar, Niki and Uszkoreit, Jakob and Jones, Llion and Gomez, Aidan N and Kaiser, {\L}ukasz and Polosukhin, Illia},
	journal={Advances in neural information processing systems},
	volume={30},
	year={2017}
}

@inproceedings{gu2024mamba,
	title={Mamba: Linear-time sequence modeling with selective state spaces},
	author={Gu, Albert and Dao, Tri},
	booktitle={First conference on language modeling},
	year={2024}
}

@article{baid2021rsna,
	title={The rsna-asnr-miccai brats 2021 benchmark on brain tumor segmentation and radiogenomic classification},
	author={Baid, Ujjwal and Ghodasara, Satyam and Mohan, Suyash and Bilello, Michel and Calabrese, Evan and Colak, Errol and Farahani, Keyvan and Kalpathy-Cramer, Jayashree and Kitamura, Felipe C and Pati, Sarthak and others},
	journal={arXiv preprint arXiv:2107.02314},
	year={2021}
}

@article{li2024brain,
	title={The brain tumor segmentation (BraTS) challenge 2023: Brain MR image synthesis for tumor segmentation (BraSyn)},
	author={Li, Hongwei Bran and Conte, Gian Marco and Hu, Qingqiao and Anwar, Syed Muhammad and Kofler, Florian and Ezhov, Ivan and van Leemput, Koen and Piraud, Marie and Diaz, Maria and Cole, Byrone and others},
	journal={ArXiv},
	pages={arXiv--2305},
	year={2024}
}

@article{zhou2017deep,
	title={Deep learning of the sectional appearances of 3D CT images for anatomical structure segmentation based on an FCN voting method},
	author={Zhou, Xiangrong and Takayama, Ryosuke and Wang, Song and Hara, Takeshi and Fujita, Hiroshi},
	journal={Medical physics},
	volume={44},
	number={10},
	pages={5221--5233},
	year={2017},
	publisher={Wiley Online Library}
}

@inproceedings{milletari2016v,
	title={V-net: Fully convolutional neural networks for volumetric medical image segmentation},
	author={Milletari, Fausto and Navab, Nassir and Ahmadi, Seyed-Ahmad},
	booktitle={2016 fourth international conference on 3D vision (3DV)},
	pages={565--571},
	year={2016},
	organization={Ieee}
}

@article{qamar2020variant,
	title={A variant form of 3D-UNet for infant brain segmentation},
	author={Qamar, Saqib and Jin, Hai and Zheng, Ran and Ahmad, Parvez and Usama, Mohd},
	journal={Future Generation Computer Systems},
	volume={108},
	pages={613--623},
	year={2020},
	publisher={Elsevier}
}

@article{cheng2023semantic,
	title={Semantic segmentation of remote sensing imagery based on multiscale deformable CNN and DenseCRF},
	author={Cheng, Xiang and Lei, Hong},
	journal={Remote Sensing},
	volume={15},
	number={5},
	pages={1229},
	year={2023},
	publisher={MDPI}
}

@article{oktay2018attention,
	title={Attention u-net: Learning where to look for the pancreas},
	author={Oktay, Ozan and Schlemper, Jo and Folgoc, Loic Le and Lee, Matthew and Heinrich, Mattias and Misawa, Kazunari and Mori, Kensaku and McDonagh, Steven and Hammerla, Nils Y and Kainz, Bernhard and others},
	journal={arXiv preprint arXiv:1804.03999},
	year={2018}
}

@article{wang2026eeo,
  title={EEO-TFV: Escape-Explore Optimizer for Web-Scale Time-Series Forecasting and Vision Analysis},
  author={Wang, Hua and Lu, Jinghao and Zhang, Fan},
  journal={arXiv preprint arXiv:2602.02551},
  year={2026}
}

@inproceedings{hu2018squeeze,
	title={Squeeze-and-excitation networks},
	author={Hu, Jie and Shen, Li and Sun, Gang},
	booktitle={Proceedings of the IEEE conference on computer vision and pattern recognition},
	pages={7132--7141},
	year={2018}
}

@inproceedings{woo2018cbam,
	title={Cbam: Convolutional block attention module},
	author={Woo, Sanghyun and Park, Jongchan and Lee, Joon-Young and Kweon, In So},
	booktitle={Proceedings of the European conference on computer vision (ECCV)},
	pages={3--19},
	year={2018}
}

@inproceedings{wang2021transbts,
	title={Transbts: Multimodal brain tumor segmentation using transformer},
	author={Wang, Wenxuan and Chen, Chen and Ding, Meng and Yu, Hong and Zha, Sen and Li, Jiangyun},
	booktitle={International conference on medical image computing and computer-assisted intervention},
	pages={109--119},
	year={2021},
	organization={Springer}
}

@article{huang2025multi,
	title={Multi-Modal Brain Tumor Segmentation via 3D Multi-Scale Self-attention and Cross-attention},
	author={Huang, Yonghao and Chen, Leiting and Zhou, Chuan},
	journal={arXiv preprint arXiv:2504.09088},
	year={2025}
}

@article{isensee2021nnu,
	title={nnU-Net: a self-configuring method for deep learning-based biomedical image segmentation},
	author={Isensee, Fabian and Jaeger, Paul F and Kohl, Simon AA and Petersen, Jens and Maier-Hein, Klaus H},
	journal={Nature methods},
	volume={18},
	number={2},
	pages={203--211},
	year={2021},
	publisher={Nature Publishing Group}
}

@article{zhang2024cf,
	title={CF-DAN: Facial-expression recognition based on cross-fusion dual-attention network},
	author={Zhang, Fan and Chen, Gongguan and Wang, Hua and Zhang, Caiming},
	journal={Computational Visual Media},
	volume={10},
	number={3},
	pages={593--608},
	year={2024},
	publisher={TUP}
}

@article{zhang2023multi,
	title={Multi-scale video super-resolution transformer with polynomial approximation},
	author={Zhang, Fan and Chen, Gongguan and Wang, Hua and Li, Jinjiang and Zhang, Caiming},
	journal={IEEE Transactions on Circuits and Systems for Video Technology},
	volume={33},
	number={9},
	pages={4496--4506},
	year={2023},
	publisher={IEEE}
}

@article{wang2024computing,
	title={Computing nodes for plane data points by constructing cubic polynomial with constraints},
	author={Wang, Hua and Zhang, Fan},
	journal={Computer Aided Geometric Design},
	volume={111},
	pages={102308},
	year={2024},
	publisher={Elsevier}
}

@article{pan2025vcanet,
	title={VcaNet: Vision Transformer with fusion channel and spatial attention module for 3D brain tumor segmentation},
	author={Pan, Dichao and Shen, Jianguo and Al-Huda, Zaid and Al-Qaness, Mohammed AA},
	journal={Computers in Biology and Medicine},
	volume={186},
	pages={109662},
	year={2025},
	publisher={Elsevier}
}

@article{zhou2025multi,
	title={Multi-level channel-spatial attention and light-weight scale-fusion network (MCSLF-Net): multi-level channel-spatial attention and light-weight scale-fusion transformer for 3D brain tumor segmentation},
	author={Zhou, Mingzhe and Li, Jinbao and Guo, Yahong},
	journal={Quantitative Imaging in Medicine and Surgery},
	volume={15},
	number={7},
	pages={6301--6325},
	year={2025},
	publisher={LWW}
}

@article{chen2023transattunet,
	title={Transattunet: Multi-level attention-guided u-net with transformer for medical image segmentation},
	author={Chen, Bingzhi and Liu, Yishu and Zhang, Zheng and Lu, Guangming and Kong, Adams Wai Kin},
	journal={IEEE Transactions on Emerging Topics in Computational Intelligence},
	volume={8},
	number={1},
	pages={55--68},
	year={2023},
	publisher={IEEE}
}

@article{liu2024vmamba,
	title={Vmamba: Visual state space model},
	author={Liu, Yue and Tian, Yunjie and Zhao, Yuzhong and Yu, Hongtian and Xie, Lingxi and Wang, Yaowei and Ye, Qixiang and Jiao, Jianbin and Liu, Yunfan},
	journal={Advances in neural information processing systems},
	volume={37},
	pages={103031--103063},
	year={2024}
}

@article{ma2024u,
	title={U-mamba: Enhancing long-range dependency for biomedical image segmentation},
	author={Ma, Jun and Li, Feifei and Wang, Bo},
	journal={arXiv preprint arXiv:2401.04722},
	year={2024}
}

@inproceedings{xing2024segmamba,
	title={Segmamba: Long-range sequential modeling mamba for 3d medical image segmentation},
	author={Xing, Zhaohu and Ye, Tian and Yang, Yijun and Liu, Guang and Zhu, Lei},
	booktitle={International conference on medical image computing and computer-assisted intervention},
	pages={578--588},
	year={2024},
	organization={Springer}
}

@article{zhou2021nnformer,
	title={nnformer: Interleaved transformer for volumetric segmentation},
	author={Zhou, Hong-Yu and Guo, Jiansen and Zhang, Yinghao and Yu, Lequan and Wang, Liansheng and Yu, Yizhou},
	journal={arXiv preprint arXiv:2109.03201},
	year={2021}
}

@inproceedings{zhang2025incomplete,
	title={Incomplete Multi-modal Brain Tumor Segmentation via Learnable Sorting State Space Model},
	author={Zhang, Zheyu and Lu, Yayuan and Ma, Feipeng and Zhang, Yueyi and Yue, Huanjing and Sun, Xiaoyan},
	booktitle={Proceedings of the Computer Vision and Pattern Recognition Conference},
	pages={25982--25992},
	year={2025}
}

@article{ali2025drbd,
	title={DRBD-Mamba for Robust and Efficient Brain Tumor Segmentation with Analytical Insights},
	author={Ali, Danish and Mian, Ajmal and Akhtar, Naveed and Hassan, Ghulam Mubashar},
	journal={arXiv preprint arXiv:2510.14383},
	year={2025}
}

@article{liu2025consistency,
	title={Consistency-Driven State-Space Model for Incomplete Multimodal MRI Brain Tumor Segmentation},
	author={Liu, Debao and Zhang, Xiaozhi and Zhou, Hong and Teo, Kok Lay},
	journal={Meta-Radiology},
	pages={100185},
	year={2025},
	publisher={Elsevier}
}

@inproceedings{zhang2026decoding,
  title={Decoding with structured awareness: integrating directional, frequency-spatial, and structural attention for medical image segmentation},
  author={Zhang, Fan and Gu, Zhiwei and Wang, Hua},
  booktitle={Proceedings of the AAAI Conference on Artificial Intelligence},
  volume={40},
  number={15},
  pages={12421--12429},
  year={2026}
}

@inproceedings{zhang2024light,
	title={Light-UNet: An Efficient Segmentation Network for Medical Image},
	author={Zhang, Yue and Xu, Chao and Zhang, Zhifan and Wang, Jianjun},
	booktitle={International Conference on Intelligent Computing},
	pages={302--313},
	year={2024},
	organization={Springer}
}

@article{xing2025segmamba,
	title={Segmamba-v2: Long-range sequential modeling mamba for general 3d medical image segmentation},
	author={Xing, Zhaohu and Ye, Tian and Yang, Yijun and Cai, Du and Gai, Baowen and Wu, Xiao-Jian and Gao, Feng and Zhu, Lei},
	journal={IEEE Transactions on Medical Imaging},
	year={2025},
	publisher={IEEE}
}

@inproceedings{ReTrack,
  title={ReTrack: Evidence-Driven Dual-Stream Directional Anchor Calibration Network for Composed Video Retrieval},
  author={Li, Zixu and Hu, Yupeng and Chen, Zhiwei and Huang, Qinlei and Qiu, Guozhi and Fu, Zhiheng and Liu, Meng},
  booktitle={Proceedings of the AAAI Conference on Artificial Intelligence},
  volume={40},
  number={28},
  pages={23373--23381},
  year={2026}
}

@inproceedings{HABIT,
  title={HABIT: Chrono-Synergia Robust Progressive Learning Framework for Composed Image Retrieval},
  author={Li, Zixu and Hu, Yupeng and Chen, Zhiwei and Zhang, Shiqi and Huang, Qinlei and Fu, Zhiheng and Wei, Yinwei},
  booktitle={Proceedings of the AAAI Conference on Artificial Intelligence},
  volume={40},
  number={8},
  pages={6762--6770},
  year={2026}
}

@inproceedings{ENCODER,
  title={Encoder: Entity mining and modification relation binding for composed image retrieval},
  author={Li, Zixu and Chen, Zhiwei and Wen, Haokun and Fu, Zhiheng and Hu, Yupeng and Guan, Weili},
  booktitle={Proceedings of the AAAI Conference on Artificial Intelligence},
  volume={39},
  number={5},
  pages={5101--5109},
  year={2025}
}

@article{FineCIR,
  title={FineCIR: Explicit Parsing of Fine-Grained Modification Semantics for Composed Image Retrieval},
  author={Li, Zixu and Fu, Zhiheng and Hu, Yupeng and Chen, Zhiwei and Wen, Haokun and Nie, Liqiang},
  journal={https://arxiv.org/abs/2503.21309},
  year={2025}
}

@inproceedings{OFFSET, 
  title = {OFFSET: Segmentation-based Focus Shift Revision for Composed Image Retrieval}, 
  author = {Chen, Zhiwei and Hu, Yupeng and Li, Zixu and Fu, Zhiheng and Song, Xuemeng and Nie, Liqiang}, 
  booktitle = {Proceedings of the ACM International Conference on Multimedia}, 
  pages = {6113–6122}, 
  year = {2025}
}

@inproceedings{HUD, 
  title = {HUD: Hierarchical Uncertainty-Aware Disambiguation Network for Composed Video Retrieval}, 
  author = {Chen, Zhiwei and Hu, Yupeng and Li, Zixu and Fu, Zhiheng and Wen, Haokun and Guan, Weili}, 
  booktitle = {Proceedings of the ACM International Conference on Multimedia}, 
  pages = {6143–6152}, 
  year = {2025} 
}

@inproceedings{INTENT,
  title={INTENT: Invariance and Discrimination-aware Noise Mitigation for Robust Composed Image Retrieval},
  author={Chen, Zhiwei and Hu, Yupeng and Fu, Zhiheng and Li, Zixu and Huang, Jiale and Huang, Qinlei and Wei, Yinwei},
  booktitle={Proceedings of the AAAI Conference on Artificial Intelligence},
  volume={40},
  number={25},
  pages={20463--20471},
  year={2026}
}

@article{REFINE,
  title={REFINE: Composed Video Retrieval via Shared and Differential Semantics Enhancement},
  author={Hu, Yupeng and Li, Zixu and Chen, Zhiwei and Huang, Qinlei and Fu, Zhiheng and Xu, Mingzhu and Nie, Liqiang},
  journal={ACM Transactions on Multimedia Computing, Communications and Applications},
  year={2026},
  publisher={ACM New York, NY}
}

@inproceedings{xiao2026points,
  title={From Points to Coalitions: Hierarchical Contrastive Shapley Values for Prioritizing Data Samples},
  author={Xiao, Canran and Dou, Jiabao and Lin, Zhiming and Ke, Zong and Hou, Liwei},
  booktitle={Proceedings of the AAAI Conference on Artificial Intelligence},
  volume={40},
  number={19},
  pages={15995--16003},
  year={2026}
}

@inproceedings{xiaoreversible,
  title={Reversible Primitive--Composition Alignment for Continual Vision--Language Learning},
  author={Xiao, Canran and Xu, Tianxiang and Jiang, Yiyang and Gao, Haoyu and Wu, Yuhan and others},
  booktitle={The Fourteenth International Conference on Learning Representations},
 year={2026}
}
}


\end{document}